\documentclass[conference]{IEEEtran}
\usepackage{booktabs}
\usepackage{graphicx}
\usepackage{amsfonts}
\usepackage{amsmath}
\usepackage{multirow}
\usepackage{marginnote}
\usepackage{xcolor}
\usepackage{url}
\usepackage{hyperref}
\hypersetup{
    colorlinks=true,
    urlbordercolor=blue,
    linkbordercolor=blue,
    linkcolor=blue,
    citecolor=red,
    urlcolor=blue,
}
\let\tss\textsuperscript
\begin{document}
\title{%
ICDAR2019 Robust Reading Challenge on Multi-lingual Scene Text Detection and Recognition -- RRC-MLT-2019
}

\author{\IEEEauthorblockN{
Nibal Nayef\tss{$\odot$1}, Yash Patel\tss{$\diamond, \circ$1}, Michal Busta\tss{$\circ$}, Pinaki Nath Chowdhury\tss{$\ddagger$}, Dimosthenis Karatzas\tss{$\bullet$},\\
Wafa Khlif\tss{$\star$}, Jiri Matas\tss{$\circ$}, Umapada Pal\tss{$\ddagger$}, Jean-Christophe Burie\tss{$\star$}, Cheng-lin Liu\tss{$\dagger$} and Jean-Marc Ogier\tss{$\star$}
}
\IEEEauthorblockA{%
$\star$: L3i Laboratory, University of La Rochelle, France\\
$\bullet$: Computer Vision Center, Universitat Aut\`{o}noma de Barcelona, Spain\\
$\ddagger$: CVPR unit, Indian Statistical Institute, India\\
$\diamond$: The Robotics Institute, Carnegie Mellon Universiry, Pittsburgh, USA\\
$\circ$: Center for Machine Perception, Department of Cybernetics, Czech Technical University, Prague, Czech Republic\\
$\dagger$: National Laboratory of Pattern Recognition, Institute of Automation of Chinese Academy of Sciences, China\\
$\odot$: Corresponding author: n.nayef@gmail.com\\[0.25em]
\small{\tss{1}: authors contributed equally as first authors}
}}
\maketitle

\begin{abstract}%
With the growing cosmopolitan culture of modern cities, the need of robust Multi-Lingual scene Text (MLT) detection and recognition systems has never been more immense. With the goal to systematically benchmark and push the state-of-the-art forward, the proposed competition builds on top of the RRC-MLT-2017 with an additional end-to-end task, an additional language in the real images dataset, a large scale multi-lingual synthetic dataset to assist the training, and a baseline End-to-End recognition method.
%
%

The real dataset consists of 20,000 images containing text from 10 languages. The challenge has 4 tasks covering various aspects of multi-lingual scene text: (a) text detection, (b) cropped word script classification, (c) joint text detection and script classification and (d) end-to-end detection and recognition. In total, the competition received 60 submissions from the research and industrial communities. This paper presents the dataset, the tasks and the findings of the presented RRC-MLT-2019 challenge.
\end{abstract}


\section{Introduction and Related Work}\label{SecIntro}
Reading scene text in natural scene images is a key component in a diverse set of applications ranging from helping the visually impaired, to data mining of street-view-like images for information used in map services and geographic information systems. Scene text detection and recognition also finds its use in larger integrated systems such as those for autonomous driving, indoor navigation and visual search engines.

This proposed competition is an extension of the RRC-MLT proposed in ICDAR-2017 \cite{Nayef17}, which was the first competition to offer the challenge of detecting multi-lingual scene text, and identifying the different scripts of such text. This proposed competition named “RRC-MLT-2019” offers the following novel aspects: 1) a new challenging task: End-to-End multi-lingual text detection and recognition 2) a synthetic dataset that matches and complements the real one in order to provide more training data 3) an additional language in the real dataset (Devanagari) 4) re-opening the 3 tasks of RRC-MLT-2017 on the new version of the real dataset and 5) an End-to-End baseline method for the new recognition task.

Research on scene text detection and recognition has primarily focused on English text, which has wide range of available datasets and well defined benchmarks \cite{CoCo,Karatzas15,Karatzas13,Gomez17}. Some other uni-lingual datasets focus on Arabic \cite{jain2017unconstrained} or French \cite{FSNS}.
The available datasets which could be considered multi-lingual have been built either for Indian languages only as in \cite{IndicSceneText2017}, or they contain only 2 scripts as in DOST \cite{Dost}, ICPR-MTWI \cite{ICPRMTWI}, MSRA-TD500\footnote{\url{http://www.iapr-tc11.org/mediawiki/index.php?title=MSRA_Text_Detection_500_Database_(MSRA-TD500)}} and KAIST\footnote{\url{http://www.iapr-tc11.org/mediawiki/index.php?title=KAIST_Scene_Text_Database}} datasets.
There exists many script identification datasets \cite{Shi16,Sharma15,Singh16,Gomez16} containing cropped scene text words from multiple languages, however, these have a relatively small number of images and their goal is limited to script classification in cropped images.

Unlike the above-mentioned datasets, we present a set of $20,000$ natural scene images containing text instances from {\sc Arabic}, {\sc Bangla}, {\sc Chinese}, {\sc Devanagari}, {\sc English}, {\sc French}, {\sc German}, {\sc Italian}, {\sc Japanese} and {\sc Korean}. We also provide a set of $277,000$ synthetically generated images with the same set of languages to assist the training. The objective of this competition is to promote the development of new methods for multi-lingual scene text understanding. The competition not only provides a large scale dataset, but also sets the evaluation protocols and standard benchmarks to promote future research.

This paper is organized as follows. Firstly, the organization of the MLT2019 challenge is outlined in Section \ref{sec:org}. The datasets used for the 4 tasks are described in Section \ref{sec:dataset}. Each task is then detailed in a separate section which contains the task's description, its evaluation protocol, the list of participant methods and their obtained results (Sections~\ref{sec:task1} to \ref{sec:task4}). We conclude the paper and discuss future work in Section~\ref{SecConclusion}. 

Due to space limitations, participant methods are listed only by their names in the tables of results, where each name is a clickable link to the details of the method (authors, affiliations, description and results). Only the winning methods for each task are described in more detail.
\section{MLT-2019 Challenge Organization}\label{sec:org}
This challenge is comprised of four tasks related to text detection (Section~\ref{sec:task1}), script classification (Section~\ref{sec:task2}), joint text detection and script identification (Section~\ref{sec:task3}) and End-to-End text recognition (Section~\ref{sec:task4}). The first three tasks have been re-opened from MLT-2017 \cite{Nayef17} on the extended MLT-2019 dataset, while the forth is a newly introduced task. The datasets created for this challenge can be relied upon to train participant methods. However, we have allowed the participants to use any other dataset to improve the training of their methods.

The web portal of the RRC platform (\url{http://rrc.cvc.uab.es/}) \cite{Karatzas15} was used for interacting with participants regarding the challenge information, schedule, downloads, online submissions and results viewing.
Overall, we had 60 different submissions distributed as follows: 25 in Task-1, 15 in Task-2, 10 in Task-3 and 10 submissions in Task-4. Some participants submitted results for more than one task, and some participants have submitted multiple \textbf{\textit{similar}} methods for the \textbf{\textit{same}} task. In the cases where the submitted methods are not demonstrably different (i.e. reflecting only parameter tuning), the participants have been asked to choose one method -- without knowing the results -- as a final submission.
\section{The ``RRC-MLT-2019" Datasets}\label{sec:dataset}
We have created two datasets: 1) The real MLT-2019 dataset that contains 20,000 real natural scene images with embedded text in 10 languages, 2) The synthetic MLT-2019 dataset that is prepared as an assistive training set only for Task-4. The synthetic dataset matches the scripts of real one.
\subsection{The MLT-2019 Dataset of Real Images}\label{sec:realdataset}
\subsubsection{Type/source of images}
The images of the dataset are natural scene images with embedded text, such as street signs, street advertisement boards, shops names, passing vehicles and users photos in microblogs. The images were captured using different mobile phone cameras or were collected from freely available images from the Internet. 
The images mainly contain intentional -- i.e. focused -- scene text, however, some unintentional text may appear in some images. Such text -- usually very small, blurry and/or occluded -- is marked to be ignored in the evaluation.
We have imposed conditions on the collection of our dataset related to the type (example: natural scenes), content (example: mostly focused text) and capture conditions of the images (example: no dark image). This is to ensure -- to some extent -- the homogeneity of the collected images as they have been collected by different people and in different countries.
\subsubsection{Number of Images, Languages and Scripts}
The dataset is comprised of 20,000 images containing text of 10 different languages (2,000 images per language). Most images contain text of more than one language, but each language is represented in at least 2,000 images.
The ten languages are: Arabic, Bangla, Chinese, Devanagari, English, French, German, Italian, Japanese and Korean.
Those languages belong to one of the following seven scripts: Arabic, Bangla, Chinese, Hindi, Japanese, Korean and Latin. An eighth script class named ``Symbols" was added for characters such as \texttt{+ / > :) ' . " -} when they appear alone in a word (without any other alphabet characters of the languages). We also have a rare script class named ``Mixed" used when characters of two or more scripts appear in the same word (without spaces).
The images are divided as follows: 50\% for training (a total of 10,000 images, 1,000 per language), and 50\% for testing.
\subsubsection{Ground Truth (GT)} 
The text in the scene images of the dataset is annotated at \textbf{\textit{word}} level. A GT-word is defined as a consecutive set of characters without spaces, i.e. words are separated by spaces, except in Chinese and Japanese where the text is labeled at line level. Each GT-word is labeled by a 4-corner bounding box, and is associated with a script class and a Unicode transcription of that GT-word.
Some text regions in the images are not readable to the annotators due to low resolution and/or other distortions. Such regions are marked as \textbf{\textit{``don't care''}} and ignored in the evaluation process.
\subsection{Synthetic Multi-Language in Natural Scene Dataset}
\label{sec:synthetic_dataset}
State-of-the-art scene text systems employ deep learning techniques which require a tremendous amount of labelled data. Hence, we have provided an additional synthetic dataset \cite{buvsta2018e2e} to complement the real one for training purposes. 
We adapt the framework proposed by Gupta \textit{et al.}\cite{gupta2016synthetic} to a multi-language setup. The framework generates realistic images by overlaying synthetic text over existing natural background images and it accounts for $3$D scene geometry.

Gupta \textit{et al.} \cite{gupta2016synthetic} proposed the following approach for scene-text image synthesis:
\begin{itemize}
\item Text in real-world usually appears in well-defined regions, which can be characterized by uniform color and texture. This is achieved by thresholding gPb-UCM contour hierarchies \cite{arbelaez2010contour} using efficient graph-cut implementation \cite{arbelaez2014multiscale}. This gives us prospective segmented regions for rendering text.
\item Dense depth map of segmented regions is then obtained using \cite{liu2015semantic} and then planer facet are fitted to them using RANSAC \cite{fischler1981random}. This way, normals to prospective regions for text rendering are estimated.
\item Finally, the text is aligned to a prospective image region for rendering. This is achieved by warping the image region to frontal-parallel view using the estimated region normals. Then, a rectangle is fitted to this region and the text is then aligned to the larger side of this rectangle.
\end{itemize}

Note that the pipeline presented in \cite{gupta2016synthetic} renders text character by character, which breaks down the ligature of {\sc Arabic}, {\sc Bangla} and {\sc Devanagari} words. We have made appropriate changes to handle this issue.

The generated dataset contains the same set of script classes as the real dataset: {\sc Arabic, Bangla, Chinese, Devanagari, Japanese, Korean, Latin}. The \textbf{Synthetic Multi-Language in Natural Scene Dataset} contains text rendered over natural scene images selected from the set of $8,000$ background images collected by \cite{gupta2016synthetic}. Annotations include word level and character level text bounding boxes along with the corresponding transcription and language class. The dataset has $277,000$ images with thousands of images for each language.
\section{Task-1: Multi-Lingual Text Detection}\label{sec:task1}
\subsection{Task-1 Description}
The objective of this task is the correct localization of multi-lingual text at word level in an image. The training set consists of 10,000 scene images, where each image has a corresponding GT file that contains a list of bounding boxes coordinates for each text word in the image. Bounding boxes are represented by four corner points ordered clock-wise. The test set has 10,000 images. For each image in the test set, participants are expected to produce a list comprised of four-corner bounding boxes for each word detected in the image.
This task was introduced in RRC-MLT-2017 \cite{Nayef17}, and it has been reopened in RRC-MLT-2019 on the 10-languages dataset.
\subsection{Evaluation Protocol for Task-1}
The f-measure (Hmean) is used as the metric for ranking the participants methods. The standard f-measure is based on both the recall and precision of the detected word bounding boxes as compared to the ground truth in all the test images (the boxes are matched/processed image by image). A detection is counted as correct (true positive) if the detected bounding box has more than 50\% overlap (intersection over union) with the GT box. At the image level, the evaluation procedure works as follows:
let $D = \{ d_1, d_2, \ldots, d_k, \ldots, d_l\}$ be the set of bounding boxes of the ``don't care'' regions,
    $G = \{ g_1, g_2, \ldots, g_i, \ldots, g_m\}$ be the set of bounding boxes in the ground truth,
and $T = \{ t_1, t_2, \ldots, t_j, \ldots, t_n \}$ be the set of bounding boxes in the results under evaluation.

First, the result bounding boxes from $T$ are matched against the ``don't care'' regions set $D$ to eliminate noise.
Each quadrilateral $t_j$ is compared against each quadrilateral $d_k$ and $t_j$ is discarded if the following condition is true:
\begin{equation}
\label{eq:task1-filter}
A(d_k) = 0 \lor \frac{A(d_k) \cap A(t_k)}{A(d_k)} > 0.5
\end{equation}
where $A(x)$ is the area of a quadrilateral $x$.
Such approach leads to some minor issues with ground truth regions overlapping with ``don't care" regions. However, only few cases in the dataset were observed, and there was no impact on the global evaluation of the methods. This highlights possible improvement of the RRC evaluation methods \cite{Karatzas13} in the future.

Once the set of the detected bounding boxes $T$ is filtered, the resulting filtered set $T'= \{ t'_1, t'_2, \ldots, t'_j, \ldots, t'_{n'} \}$ is matched against the set of ground truth quadrilaterals $G$.
A positive match is counted each time a couple of elements $(g_i, t'_j)$ verifies the following condition:
\begin{equation}
\label{eq:task1-positive-match}
\frac{A(g_i) \cap A(t'_j)}{A(g_i) \cup A(t'_j)} > 0.5
\end{equation}
with $g_i \in G$ and $t'_j \in T'$.
An extra test ensures that each element $g_i$ and each element $t'_j$ can only be matched once.

At the whole test set level, the evaluation metrics are computed cumulatively from all the test images (detection results of all the images are pooled together). Extending the set of positive (relevant) matches $M$, the set of expected words $G$ and the set of filtered results $T'$ to include all the test images, we can compute the precision, recall and f-measure as follows:
\begin{equation}
\label{eq:task1-prec-rec-fmes}
\begin{array}{c}
\text{precision} = \frac{|M|}{|T'|}\\[0.5em]
\text{recall} = \frac{|M|}{|G|}\\[0.5em]
\text{f-measure} = \frac{2 \cdot \text{precision} \cdot \text{recall}}{\text{precision} + \text{recall}}
\end{array}
\end{equation}
\subsection{Participant Methods and Results for Task-1}\label{PT1}
We report here the results obtained by the participants for this task. 
The ranking of the participants according to Hmean is summarized in Table~\ref{tab:task1-results}. The name of each participant method in the table is a link to its online description and results.
\begin{table}[tb]
\caption{Results of the RRC-MLT-2019 Challenge for Task-1: Multi-Lingual Text Detection}
\label{tab:task1-results}
\begin{center}
\begin{tabular}{rlllr}
\toprule
\textbf{Rank} & \textbf{Method} & \textbf{Hmean} & \textbf{Precision} & \textbf{Recall}\\
\midrule %
1 & \href{https://rrc.cvc.uab.es/?ch=15&com=evaluation&view=method_info&task=1&m=57839}{Tencent-DPPR Team} & 83.61\% & 87.52\%	& 80.05\% \\%
1 & \href{https://rrc.cvc.uab.es/?ch=15&com=evaluation&view=method_info&task=1&m=57864}{Multi-stage\_Text\_Detector} & 83.59\%	& 87.75\% & 79.80\% \\%
2 & \href{https://rrc.cvc.uab.es/?ch=15&com=evaluation&view=method_info&task=1&m=57521}{NJU-ImagineLab} & 83.07\%	& 87.85\%	& 78.79\%\\%
3 & \href{https://rrc.cvc.uab.es/?ch=15&com=evaluation&view=method_info&task=1&m=56923}{PMTD} \cite{liu2019pyramid} & 82.53\% & 87.47\% & 78.12\%\\%
4 & \href{https://rrc.cvc.uab.es/?ch=15&com=evaluation&view=method_info&task=1&m=56726}{MaskRCNN$++$} & 80.35\% & 82.64\% & 78.19\%\\%
5 & \href{https://rrc.cvc.uab.es/?ch=15&com=evaluation&view=method_info&task=1&m=56717}{IC\_RL} & 80.11\% & 82.97\% & 77.44\%\\%
6 & \href{https://rrc.cvc.uab.es/?ch=15&com=evaluation&view=method_info&task=1&m=56700}{4Paradigm-Data-Intelligence} & 79.84\% & 83.44\% & 76.54\%\\%
\multirow{2}{*}{7} & \href{https://rrc.cvc.uab.es/?ch=15&com=evaluation&view=method_info&task=1&m=57333}{Two-stage Text Detector} & \multirow{2}{*}{78.38\%} & \multirow{2}{*}{82.26\%} & \multirow{2}{*}{74.85\%}\\%
 & \href{https://rrc.cvc.uab.es/?ch=15&com=evaluation&view=method_info&task=1&m=57333}{---based on Cascade-RCNN}\\
8 & \href{https://rrc.cvc.uab.es/?ch=15&com=evaluation&view=method_info&task=1&m=57704}{MM-MaskRCNN} & 76.79\%	& 84.73\%	& 70.21\% \\
9 & \href{https://rrc.cvc.uab.es/?ch=15&com=evaluation&view=method_info&task=1&m=57762}{TH-DL} & 76.64\%	& 84.55\%	& 70.09\% \\
10 & \href{https://rrc.cvc.uab.es/?ch=15&com=evaluation&view=method_info&task=1&m=57613}{SOT} & 74.24\%	& 79.96\%	& 69.28\%\\%
11 & \href{https://rrc.cvc.uab.es/?ch=15&com=evaluation&view=method_info&task=1&m=57257}{DISTILLED CRAFT} & 72.94\%	& 81.22\%	& 66.19\% \\
12 & \href{https://rrc.cvc.uab.es/?ch=15&com=evaluation&view=method_info&task=1&m=57702}{Text-Mountain} & 71.95\%	& 72.12\%	& 71.77\% \\
13 & \href{https://rrc.cvc.uab.es/?ch=15&com=evaluation&view=method_info&task=1&m=57473}{CRAFTS} \cite{baek2019character} & 70.86\%	& 81.42\%	& 62.73\% \\%
\multirow{2}{*}{14} & \href{https://rrc.cvc.uab.es/?ch=15&com=evaluation&view=method_info&task=1&m=57736}{Unicamp-SRBR-} & \multirow{2}{*}{70.81\%} & \multirow{2}{*}{81.58\%} &\multirow{2}{*}{62.54\%}\\
 & \href{https://rrc.cvc.uab.es/?ch=15&com=evaluation&view=method_info&task=1&m=57736}{---MLT2019-PELEEText} & & &\\
15 & \href{https://rrc.cvc.uab.es/?ch=15&com=evaluation&view=method_info&task=1&m=57674}{RRPN} & 69.56\% & 77.71\%	& 62.95\% \\
\multirow{2}{*}{16} & \href{https://rrc.cvc.uab.es/?ch=15&com=evaluation&view=method_info&task=1&m=57735}{Unicamp-SRBR-MLT2019} & \multirow{2}{*}{68.56\%} & \multirow{2}{*}{77.00\%} & \multirow{2}{*}{61.79\%}\\
 & \href{https://rrc.cvc.uab.es/?ch=15&com=evaluation&view=method_info&task=1&m=57735}{---Fusion-PSENet-PELEEText} &  & & \\
17 & \href{https://rrc.cvc.uab.es/?ch=15&com=evaluation&view=method_info&task=1&m=57850}{Lomin OCR} & 67.65\%	& 71.62\%& 64.09\% \\
18 & \href{https://rrc.cvc.uab.es/?ch=15&com=evaluation&view=method_info&task=1&m=57484}{NXB OCR} & 65.96\%	& 70.59\%	& 61.90\% \\
19 & \href{https://rrc.cvc.uab.es/?ch=15&com=evaluation&view=method_info&task=1&m=56236}{PSENet} & 65.83\% & 73.52\% & 59.59\%\\%
20 & \href{https://rrc.cvc.uab.es/?ch=15&com=evaluation&view=method_info&task=1&m=56441}{MLT2019 ETD} & 64.36\% & 78.71\% & 54.44\%\\%
21 & \href{https://rrc.cvc.uab.es/?ch=15&com=evaluation&view=method_info&task=1&m=56567}{CLTDR} & 63.53\% & 77.20\% & 53.97\%\\%
22 & \href{https://rrc.cvc.uab.es/?ch=15&com=evaluation&view=method_info&task=1&m=57610}{TP} & 58.01\%	& 77.59\%	& 46.32\%\\	
23 & \href{https://rrc.cvc.uab.es/?ch=15&com=evaluation&view=method_info&task=1&m=56556}{based on mask rcnn} & 49.45\% & 64.69\% & 40.02\%\\%
24 & \href{https://rrc.cvc.uab.es/?ch=15&com=evaluation&view=method_info&task=1&m=57729}{Cyberspace} & 47.09\%	& 69.48\%	& 35.61\% \\
\bottomrule
\end{tabular}
\end{center}
\end{table}

Most of the participant methods -- including winner methods -- are based on R-CNN (masked, cascaded, with refinement stage etc.). This shows that the R-CNN method can be improved to achieve highly accurate detection. Other methods have used one or more deep nets used previously for text detection and recognition such as ResNet, EAST (based on FCN), RRPN and FPN among others.
\subsubsection{Winner Methods of Task-1}\label{winnerT1}\hfill\\
We have two winner methods for this task (both ranked 1). The difference in Hmean of the two methods is not significant given the possibility of the presence of some errors in the GT.
The first winner method is called \textbf{``Tencent-DPPR Team"}.
\\\textbf{Authors:} Longhuang Wu, Shangxuan Tian, Chang Liu, Wenjie Cai, Jiachen Li, Sicong Liu, Haoxi Li, Chunchao Guo, Hongfa Wang, Hongkai Chen, Qinglin Lu, Chun Yang, Xucheng Yin, Lei Xiao.
\\\textbf{Affiliation:} Tencent-DPPR (Data Platform Precision Recommendation) team.
\\\textbf{Method description:} The text detector follows the framework of Mask R-CNN that employs a mask to detect multi-oriented scene texts. The text detector is trained using the RRC-MLT-2019 training set and the MSRA-TD500 dataset. A multi-scale training approach is used during training. To obtain the final ensemble results, two different backbones and different multi-scale testing approaches are combined.

The other winner method is \textbf{``Multi-stage Text Detector"}.
\\\textbf{Authors:} Pengfei Wang\tss{$\ddagger, \circ$}, Mengyi En\tss{$\circ$}, Xiaoqiang Zhang\tss{$\circ$}, Chengqaun Zhang\tss{$\circ$}.
\\\textbf{Affiliation:} VIS-VAR Team at Baidu Inc.\tss{$\circ$} and Xidian University\tss{$\ddagger$}. Pengfei Wang did carried out this work while interning at Baidu Inc.
\\\textbf{Method description:} The method relies on two stages. The first stage is a modified Mask-R-CNN, where a rotated proposal module is introduced to make Mask-RCNN more suitable for detecting multi-oriented scene text. The second stage is a refinement to get the final detection results.
\section{Task-2: Cropped Word Script identification}\label{sec:task2}
\subsection{Task Description}
The objective of this task is to identify the script of a cropped word image. The training and test sets of this task consist of cropped word images that have been extracted from the full scene images of Task-1 based on the bounding boxes of the GT words. In total, there are 89,177 training images and 102,462 test images.
The text in our dataset images appears in 10 different languages, some of them share the same script. Additionally, punctuation and some math symbols sometimes appear as separate words, those words are annotated as a special script class called ``Symbols". Hence, we have a total of 8 different scripts. We have excluded the words that have ``Mixed" script for this task due to the very small number of samples. We have also excluded all the ``don't care" words whether they have a recognizable script or not.

Given the test images of cropped words, participants are asked to identify the script \textit{ID} of each word image file. A single script name (\textit{ID}) per image is requested. The valid scripts for this task are: ``Arabic", ``Bangla", ``Chinese", ``Hindi", ``Japanese", ``Korean", ``Latin" and ``Symbols". This task was introduced in RRC-MLT-2017 \cite{Nayef17}, and it has been reopened in RRC-MLT-2019 on the new dataset of 10 languages.
\subsection{Evaluation Protocol for Task-2}
The evaluation and the ranking of results is based on classification accuracy. Participants provide a script \textit{ID} for each word image, and if the result is correct, then the count of correct results is incremented. The overall accuracy of a given method is is defined as follows.
let $G = \{ g_1, g_2, \dots, g_i, \dots, g_m \}$ be the set of correct script classes in the ground truth,
and $T = \{ t_1, t_2, \dots, t_i, \dots, t_m \}$ be the set of script classes returned by a given method,
where $g_i$ and $t_i$ refer to the same original image.
Then, the performance of a given method is expressed by:
\begin{equation}
\label{eq:task2-eval}
\text{accuracy} = \frac{1}{m} \sum\limits_{i=1, \ldots, m}
 \begin{cases}
    1  & \quad \text{if }g_i = t_i\\
    0  & \quad \text{otherwise}\\
  \end{cases}
\end{equation}
\subsection{Participant Methods and Results for Task-2}\label{PT2}
We report here the results obtained by the participants for this task.
The ranking of the participants -- according to script classification accuracy -- is summarized in Table~\ref{tab:task2-results}.
\begin{table}[tb]
\caption{Results of the RRC-MLT-2019 Challenge for Task-2: Cropped Word Script identification}
\label{tab:task2-results}
\begin{center}
\begin{tabular}{rlr}
\toprule
\textbf{Rank} & \textbf{Method} & \textbf{Accuracy} \\
\midrule
1 & \href{https://rrc.cvc.uab.es/?ch=15&com=evaluation&view=method_info&task=2&m=57836}{Tencent-DPPR Team} & 94.03\% \\
2 & \href{https://rrc.cvc.uab.es/?ch=15&com=evaluation&view=method_info&task=2&m=57647}{SOT: CNN-based Classifier} & 91.66\% \\
3 & \href{https://rrc.cvc.uab.es/?ch=15&com=evaluation&view=method_info&task=2&m=56664}{GSPA\_HUST} & 91.02\% \\
3 & \href{https://rrc.cvc.uab.es/?ch=15&com=evaluation&view=method_info&task=2&m=57592}{SCUT-DLVC-Lab} & 90.97\% \\
4 & \href{https://rrc.cvc.uab.es/?ch=15&com=evaluation&view=method_info&task=2&m=57867}{TPS-ResNet} \cite{baek2019wrong} & 90.90\% \\
4 & \href{https://rrc.cvc.uab.es/?ch=15&com=evaluation&view=method_info&task=2&m=57821}{Conv-Transformer} & 90.88\%\\%
5 & \href{https://rrc.cvc.uab.es/?ch=15&com=evaluation&view=method_info&task=2&m=57763}{TH-DL} & 90.70\% \\
6 & \href{https://rrc.cvc.uab.es/?ch=15&com=evaluation&view=method_info&task=2&m=56608}{TH-ML} & 88.85\%\\%
7 & \href{https://rrc.cvc.uab.es/?ch=15&com=evaluation&view=method_info&task=2&m=56646}{MultiScale\_HUST} & 88.64\%\\%
8 & \href{https://rrc.cvc.uab.es/?ch=15&com=evaluation&view=method_info&task=2&m=56561}{USTC \& IFLYTEK} & 88.54\%\\%
9 & \href{https://rrc.cvc.uab.es/?ch=15&com=evaluation&view=method_info&task=2&m=57341}{Conv\_Attention} & 88.41\% \\
10 & \href{https://rrc.cvc.uab.es/?ch=15&com=evaluation&view=method_info&task=2&m=56938}{Cold} & 87.98\%\\%
11 & \href{https://rrc.cvc.uab.es/?ch=15&com=evaluation&view=method_info&task=2&m=56475}{NXB OCR} & 84.88\%\\%
12 & \href{https://rrc.cvc.uab.es/?ch=15&com=evaluation&view=method_info&task=2&m=55090}{ELE-MLT based method} & 82.86\%\\%
13 & \href{https://rrc.cvc.uab.es/?ch=15&com=evaluation&view=method_info&task=2&m=57470}{Res\_MUL\_SPP\_BUPT} & 71.31\%\\%
\bottomrule
\end{tabular}
\end{center}
\end{table}

Most of the participant methods base their methods on famous deep nets for text recognition such as ResNet, VGG16, Seq2Seq with CTC, CNN with self-attention, RNN, CNN-LSTM etc., with adding some improvements such as multi-scale techniques, attention, voting strategy for combining results from multiple nets, training statistics of the scripts etc. 
\subsubsection{Winner Method of Task-2}\label{winnerT2}\hfill\\
The winner method is called \textbf{``Tencent-DPPR Team"}.
\\\textbf{Authors:} Sicong Liu, Haoxi Li, Haibo Qin, Ben Xu, Chunchao Guo, Longhuang Wu, Shangxuan Tian, Hongfa Wang, Hongkai Chen, Qinglin Lu, Chun Yang, Xucheng Yin, Lei Xiao.
\\\textbf{Affiliation:} Tencent-DPPR (Data Platform Precision Recommendation) team.
\\\textbf{Method description:} In the first stage, the method recognizes text-lines and their character-level language types using the ensemble results of several recognition models which are based on Seq2Seq with CTC and CNN with self-attention \& RNN. In the second stage, the language types of the recognized results are identified based on the statistics of the MLT-2019 training set and the Wikipedia corpus.
\section{Task-3: Joint Text Detection and Script Identification}\label{sec:task3}
\subsection{Task Description}
The objective of this task is the correct localization of all the words in a full scene image and jointly identifying the script \textit{ID} of each localized (detected) word.
The training and test sets are comprised of 10,000 images each, the same scene images described in Task-1.
The ground truth file corresponding to an image contains the coordinates of the bounding boxes of all the words inside the image (including ``don't care" words), the transcription and the script \textit{ID} for each word box.

Participants are required to output the list of the detected bounding boxes for each image \textbf{\textit{and}} the script \textit{ID} for each detected bounding box (word) in the list. This task was introduced in RRC-MLT-2017 \cite{Nayef17}, and it has been reopened in RRC-MLT-2019 on the new dataset of 10 languages.
\subsection{Evaluation Protocol for Task-3}
The evaluation of this task is a cascade of correct localization of a text box \textbf{\textit{and}} its correct script classification. This only requires injecting the control of the correct identification of the script for a given text region into Equation~\ref{eq:task1-positive-match}:
\begin{equation}
\label{eq:task3-positive-match}
\frac{A(g_i) \cap A(t'_j)}{A(g_i) \cup A(t'_j)} > 0.5\\
\land \text{ScriptId}(t'_j) = \text{ScriptId}(g_i)
\end{equation}
Except for the definition of a correct detection, Task-3 has the same ranking and evaluation protocol as Task-1. A correct detection here is counted when the box is both detected correctly and its correct script \textit{ID} is identified.
\subsection{Participant Methods and Results for Task-3}\label{PT3}
We report here the results obtained by the participants for this joint detection and classification task.
The ranking of the participants' methods -- according to Hmean -- is summarized in Table~\ref{tab:task3-results}.
\begin{table}[tb]
\caption{Results of the RRC-MLT-2019 Challenge for Task-3: Joint Text Detection and Script Identification}
\label{tab:task3-results}
\begin{center}
\begin{tabular}{rlllr}
\toprule
\textbf{Rank} & \textbf{Method} & \textbf{Hmean} & \textbf{Precision} & \textbf{Recall}\\
\midrule
1 & \href{https://rrc.cvc.uab.es/?ch=15&com=evaluation&view=method_info&task=3&m=57862}{Tencent-DPPR Team} & 80.84\% & 87.68\% & 74.99\% \\%
2 & \href{https://rrc.cvc.uab.es/?ch=15&com=evaluation&view=method_info&task=3&m=57841}{Mask\_RCNN-transformer} & 75.12\% & 77.26\% & 73.10\%\\%
3 & \href{https://rrc.cvc.uab.es/?ch=15&com=evaluation&view=method_info&task=3&m=57575}{icdar2019\_mlt\_task3\_test\_lqj} & 72.13\% & 74.21\% & 70.16\%\\%
4 & \href{https://rrc.cvc.uab.es/?ch=15&com=evaluation&view=method_info&task=3&m=57764}{TH-DL} & 71.01\% & 78.34\% & 64.94\%\\%
5 & \href{https://rrc.cvc.uab.es/?ch=15&com=evaluation&view=method_info&task=3&m=56687}{DISTILLED CRAFT} & 68.69\% & 74.97\% & 63.39\%\\%
6 & \href{https://rrc.cvc.uab.es/?ch=15&com=evaluation&view=method_info&task=3&m=57709}{Cold} & 68.58\% & 77.79\% & 61.32\%\\%
7 & \href{https://rrc.cvc.uab.es/?ch=15&com=evaluation&view=method_info&task=3&m=57475}{CRAFTS} \cite{baek2019character} & 68.34\% & 78.52\% & 60.50\% \\%
8 & \href{https://rrc.cvc.uab.es/?ch=15&com=evaluation&view=method_info&task=3&m=57604}{SOT + Classifier} & 65.66\% & 66.20\% & 65.13\%\\%
9 & \href{https://rrc.cvc.uab.es/?ch=15&com=evaluation&view=method_info&task=3&m=57746}{USTC \& IFLYTEK: det+cls} & 63.14\% & 63.30\% & 62.98\%\\%
10 & \href{https://rrc.cvc.uab.es/?ch=15&com=evaluation&view=method_info&task=3&m=57488}{NXB OCR} & 57.74\% & 61.79\% & 54.18\%\\%
\bottomrule
\end{tabular}
\end{center}
\end{table}

Almost all the methods that participated in Task-1, have also participated in Tasks 3 and 4 as the core task of detecting text words has been accomplished in Task-1. This shows that the research has moved towards end-to-end approaches with the same underlying deep learning-based methods.
\subsubsection{Winner Method of Task-3}\hfill\\
The winner method is called \textbf{``Tencent-DPPR Team"} presented by the same team which won Tasks 1 \& 2 as well. This can be expected as Task-3 is a joint between the first two tasks. Indeed, the winner method here is a cascade of the two methods presented by the Tencent-DPPR for Tasks 1 \& 2.
\\\textbf{Authors:} Longhuang Wu, Shangxuan Tian, Haoxi Li, Sicong Liu, Jiachen Li, Chunchao Guo, Haibo Qin, Chang Liu, Hongfa Wang, Hongkai Chen, Qinglin Lu, Chun Yang, Xucheng Yin, Lei Xiao.
\\\textbf{Affiliation:} Tencent-DPPR (Data Platform Precision Recommendation) team.
\\\textbf{Method description:} A cascade of the team's two descriptions of their methods for the first two tasks (see Subsections \ref{winnerT1} and \ref{winnerT2}).
\section{Task-4: End-to-End Text Detection and Recognition}\label{sec:task4}
\subsection{Task-4 Description}
This newly introduced task is very challenging: a unified OCR for multiple-languages. We present the task of end-to-end scene text detection and recognition in multi-lingual setting that is coherent with its English-only counterparts. Given an input scene image, the objective is to localize a set of bounding boxes and their corresponding transcriptions.

The training and test sets are comprised of 10,000 images each, they are the same scene images described in Task-1 and with the same GT as in Task-3.
The training data is unbalanced for the different languages in the real dataset (see Subsection \ref{sec:realdataset}). Hence, to help with the training for this task, we provide to participants the synthetic dataset described in Subsection \ref{sec:synthetic_dataset} in addition to the real dataset. 
\subsection{Evaluation Protocol for Task-4}
The evaluation of this task is a cascade of correct localization (detection) of a text box \textbf{\textit{and}} its correct transcription. This only requires injecting the control of matching the transcription for a given text region into Equation~\ref{eq:task1-positive-match}:
\begin{equation}
\label{eq:task4-positive-match}
\frac{A(g_i) \cap A(t'_j)}{A(g_i) \cup A(t'_j)} > 0.5\\
\land \text{transcription}(t'_j) = \text{transcription}(g_i)
\end{equation}
where the transcription is matched with case insensitive setting, and a given transcription result must exactly match the GT transcription (i.e the edit distance between the two transcriptions is zero).
Except for the definition of a correct detection, Task-4 has the same ranking and evaluation protocol as Task-1. A correct detection here is counted when the box is both detected correctly and its text is recognized correctly.

Note that the test set words which contain characters that did not appear in the train set are set as ``don't care" for both the detection and recognition. This means whether a method detects them correctly or not, or recognize them correctly or not, they won't be counted in the evaluation.
\subsection{Participant Methods and Results for Task-4}\label{PT4}
We report here the results obtained by the participants for this task.
The ranking of the participants according to Hmean is summarized in Table~\ref{tab:task4-results}. Note that the online results show additional evaluation metrics including the edit-distance accuracy for recognition part.
\begin{table}[tb]
\caption{Results of the RRC-MLT-2019 Challenge for Task-4: End-to-End Text Detection and Recognition}
\label{tab:task4-results}
\begin{center}
\begin{tabular}{rlllr}
\toprule
\textbf{Rank} & \textbf{Method} & \textbf{Hmean} & \textbf{Precision} & \textbf{Recall}\\
\midrule
\multirow{2}{*}{1} & \href{https://rrc.cvc.uab.es/?ch=15&com=evaluation&view=method_info&task=4&m=57851}{Tencent-DPPR Team} & \multirow{2}{*}{59.15\%} & \multirow{2}{*}{71.26\%} & \multirow{2}{*}{50.55\%}\\%
 & \href{https://rrc.cvc.uab.es/?ch=15&com=evaluation&view=method_info&task=4&m=57851}{\& USTB-PRIR}\\
2 & \href{https://rrc.cvc.uab.es/?ch=15&com=evaluation&view=method_info&task=4&m=57696}{end2end} & 52.50\% & 55.34\% & 49.93\%\\%
3 & \href{https://rrc.cvc.uab.es/?ch=15&com=evaluation&view=method_info&task=4&m=57476}{CRAFTS} \cite{baek2019character} & 51.74\% & 65.68\% & 42.68\%\\%
4 & \href{https://rrc.cvc.uab.es/?ch=15&com=evaluation&view=method_info&task=4&m=57842}{Mask\_RCNN-transformer} & 51.04\% & 52.51\% & 49.64\% \\
5 & \href{https://rrc.cvc.uab.es/?ch=15&com=evaluation&view=method_info&task=4&m=57748}{Three-stage method} & 40.19\% & 44.37\% & 36.73\%\\
6 & \href{https://rrc.cvc.uab.es/?ch=15&com=evaluation&view=method_info&task=4&m=57706}{USTC \& IFLYTEK} & 39.55\% & 39.71\% & 39.39\%\\%
7 & \href{https://rrc.cvc.uab.es/?ch=15&com=evaluation&view=method_info&task=4&m=57565}{icdar2019\_mlt\_test\_lqj} & 38.75\% & 39.88\% & 37.67\%\\
8 & \href{https://rrc.cvc.uab.es/?ch=15&com=evaluation&view=method_info&task=4&m=57765}{TH-DL} & 37.32\% & 41.22\% & 34.10\%\\%
9 & \href{https://rrc.cvc.uab.es/?ch=15&com=evaluation&view=method_info&task=4&m=57732}{RRPN+CLTDR} & 33.82\% & 38.62\% & 30.08\%\\%
10 & \href{https://rrc.cvc.uab.es/?ch=15&com=evaluation&view=method_info&task=4&m=57493}{NXB OCR} & 32.07\% & 34.37\% & 30.06\%\\%
-- & \href{https://rrc.cvc.uab.es/?ch=15&com=evaluation&view=method_info&task=4&m=55986}{E2E-MLT ``Baseline"} \cite{buvsta2018e2e} & 26.46\% & 37.44\% & 20.47\%\\%
\bottomrule
\end{tabular}
\end{center}
\end{table}

For this new task, we present a baseline method named in Table~\ref{tab:task4-results} as ``E2E-MLT" \cite{buvsta2018e2e} with available source code\footnote{\url{https://github.com/MichalBusta/E2E-MLT/blob/master/README.md}}. Since it has been submitted by the organizers as a baseline, we do not rank it.

Once again we note that most of the methods have also participated in Tasks 1 \& 3 and incorporated a recognition part into their networks (The recognition parts have been used -- in some cases -- to help with Task-2). Participant methods mostly rely on the detection and recognition deep nets mentioned in Subsection \ref{PT1}, and adding to this list: attention-based decoders, MORAN-v2, CRNN adopting CTC, CRNN and convolutional transformers with a ResNet50 backbone. We have also noted that combining multiple nets can be effective in end-to-end tasks.
\subsubsection{Winner Method of Task-4}\hfill\\
The winner method is called \textbf{``Tencent-DPPR Team \& USTB-PRIR"} as a collaboration between two teams. The Tencent-DPPR Team has indeed won the 4 tasks of our MLT-2019 challenge. The team shared the same rank with another team in Task 1, and collaborated with another team in Task 4.
\\\textbf{Authors:} Sicong Liu, Longhuang Wu, Shangxuan Tian, Haoxi Li, Chunchao Guo, Haibo Qin, Chang Liu, Hongfa Wang, Hongkai Chen, Qinglin Lu, Chun Yang, Xucheng Yin, Lei Xiao.
\\\textbf{Affiliations:} Tencent-DPPR \& USTB-PRIR.
\\\textbf{Method description:} The detection part of the framework is the same as described in Subsection \ref{winnerT1}, and the recognition part is the same as described in Subsection \ref{winnerT2} as it was used for the script classification task of detected words.
\section{Conclusions and Future Directions}\label{SecConclusion}
This report has summarized the organization and the findings of the multi-lingual scene text (MLT) challenge of the RRC competition. There has been a total of 60 different submissions distributed over the four proposed tasks. This shows a big interest of the community in the problem of multi-lingual scene text detection and recognition. This interest has vastly grown since the 2017-edition of MLT.

Our work has extended the previous RRC-MLT-2017 edition in the following aspects: adding a new language (of a new script), introducing a new end-to-end task for text recognition and building a new synthetic dataset that matches the real one in terms of scripts for training purposes.
All the details about the RRC-MLT-2019 challenge and its datasets are available on the RRC competition website: \url{http://rrc.cvc.uab.es/?ch=15}.

Future versions of this challenge could focus on increasing the number of languages in the dataset (of similar and also of different scripts) leading to very large-scale problems in multi-lingual scene text detection and recognition. Moreover, there is a need to design more robust evaluation protocols that can handle special appearances of text such as unfocused scene text, and also deal with sub-task evaluation for ``don't care" words in joint or end-to-end tasks. This work provides the base on which such future work could be built.

\section*{Acknowledgments}
This work is partially funded by Agence Nationale de la Recherche (ANR) in France, National Natural Science Foundation of China (NSFC 61411136002) in China under the AUDINM project,
and by the Visual Computing Competence Center TE01020415 of the Technology Agency of the Czech Republic.

\bibliographystyle{IEEEtran}
\bibliography{references}
\end{document}